\documentclass{article}

\usepackage{arxiv}

\usepackage[utf8]{inputenc} 
\usepackage[T1]{fontenc}    
\usepackage{hyperref}       
\usepackage{url}            
\usepackage{booktabs}       
\usepackage{amsfonts}       
\usepackage{nicefrac}       
\usepackage{microtype}      
\usepackage{lipsum}
\usepackage{graphicx}
\usepackage{url}
\usepackage{xcolor}
\usepackage{amsmath}
\usepackage{mathtools}
\usepackage{tabularx}
\graphicspath{ {./images/} }

\title{Saliency Map-based Image Retrieval using Invariant Krawtchouk Moments}

\author{
 Ashkan Nejad \\
  Department of Computer Science\\
  Institute for Advanced Studies in Basic Sciences\\
  Zanjan, Iran\\
  \texttt{www.nejad.info} \\
   \And
 Mohammad Reza Faraji \\
  Department of Computer Science\\
  Institute for Advanced Studies in Basic Sciences\\
  Zanjan, Iran\\
  \And
 Xiaojun Qi \\
  Department of Computer Science\\
  Utah State University\\
  Logan, Utah, USA \\
}

\begin{document}
\maketitle
\begin{abstract}
With the widespread adoption of digital devices equipped with cameras and the rapid development of Internet technology, numerous content-based image retrieval systems and novel image feature extraction techniques have emerged in recent years. This paper introduces a saliency map-based image retrieval approach using invariant Krawtchouk moments (SM-IKM) to enhance retrieval speed and accuracy. The proposed method applies a global contrast-based salient region detection algorithm to create a saliency map that effectively isolates the foreground from the background. It then combines multiple orders of invariant Krawtchouk moments (IKM) with local binary patterns (LBPs) and color histograms to comprehensively represent the foreground and background. Additionally, it incorporates LBPs derived from the saliency map to improve discriminative power, facilitating more precise image differentiation. A bag-of-visual-words (BoVW) model is employed to generate a codebook for classification and discrimination. By using compact IKMs in the BoVW framework and integrating a range of region-based features—including color histograms, LBPs, and saliency map-enhanced LBPs—our proposed SM-IKM achieves efficient and accurate image retrieval. Extensive experiments on publicly available datasets, such as Caltech 101 and Wang, demonstrate that SM-IKM outperforms recent state-of-the-art retrieval methods. The source code for SM-IKM is available at \url{github.com/arnejad/SMIKM}.
 
\end{abstract}


\section{Introduction}
\label{sec1}
In the past decades, there has been a need for finding the most semantically similar samples to a query image in many applications, including medicine \cite{Choe2021}, copyright protection \cite{Wang2021} and e-commerce \cite{Baldrati2022}. Since the images cannot necessarily be found by describing them in words, it is essential to design a procedure to compare and find the most similar samples to a query based on the content. This has led to the development of content-based image retrieval (CBIR) systems \cite{Tyagi2018}, which consider various properties such as color, texture, shape, and structure  \cite{Monowar2022} to semantically compare the content of images. However, semantic analysis of images makes CBIR challenging. In addition, different images of the same object are prone to transformations (e.g., translation, rotation, and scaling). As a result, a CBIR system needs to maintain retrieval accuracy when images in the dataset are captured with various transformations \cite{Tyagi2018}.

Various techniques have been applied for the improvement of CBIR systems. One of the first methods is bag-of-visual-words (BoVW) \cite{Yang2010} ], which is derived from the bag-of-words \cite{Joachims1998} concept used in natural language processing. Researchers employ the original BoVW method and its variants to generate a codebook for classification and discrimination. They further utilize the codebook together with a feature descriptor such as SIFT \cite{Lowe1999} or SURF \cite{Bay2006} to achieve higher retrieval accuracy  \cite{Chen2014, Sarwar2019, Bai2014}.


Many studies have been conducted to propose algorithms that extract robust and discriminative features. For example, moments are constructed based on continuous or discrete polynomials to resolve image transformation issues.   Zernik and Legendre moments are examples of continuous orthogonal moments. They are valid in the [-1, 1] range and therefore suffer from the errors resulting from the mandatory transformation and quantization processes \cite{Yap2003}. On the other hand, the discrete orthogonal moments like
Tchebichef and Krawtchouk do not carry the digitization, and coordinate space transformation errors since the basis sets are orthogonal in the discrete domain \cite{Yap2003}. Unlike Tchebichef moments, Krawtchouk moments can work as local descriptors \cite{Karampasis2021}.

To further increase discrimination among images, researchers use the fusion technique to combine multiple features \cite{Sun2005, Fernando2012, Xie2020, Garg2021}, where each feature provides information from different aspects, including illumination, local pattern, and color. For example, Singh et al. \cite{Singh2018} propose an LBPC+LBPH+CH framework that acquires a fusion of local binary pattern for color images (LBPC), the local binary pattern of the hue (LBPH), and the color histogram (CH) features. The final fused features then take part in similarity measurement to compute the closest images to the query. To address the challenge of describing local texture patterns in color images, Dubey et al. \cite{Dubey2016} propose adder- and decoder-based schemas called MDLBP to combine the LBPs from more than one channel.

Since not all areas in the image contribute equally to the recognition process, researchers have recently used saliency maps to take regional importance into account. The saliency map is a matrix of the same size as the input image, where each cell in the map indicates its prominence in the image, and the main objects correspond to the pixels with high prominence. Zhang et al. \cite{Zhang2017} propose an \emph{extended salient region} (ESR) method to obtain the salient region with a surrounding safety margin. They then extract texture, SIFT, and HSV histogram features from each ESR to form a codebook to compare the content of the images.

One recent CBIR method, which combines BoVW and salient region detection, is the saliency-based multi-feature modeling (SMFM) \cite{Bai2018} method. It obtains foreground and background areas in images via the saliency map and then extracts fused features, including SIFT, LBP, and color histograms for both foreground and background regions. Integrating saliency maps in the image retrieval process makes retrieval results semantically closer to human perception.

In this paper, we propose a CBIR framework named saliency map-based image retrieval using invariant Krawtchouk moments (SM-IKM), which uses salient regions to customize the extent of contribution for each segment of an image. SM-IKM first separates the foreground and background regions of each image using the global contrast-based salient region detection method \cite{Cheng2014}. It then exploits invariant Krawtchouk moments (IKMs) to gain transformation robustness and more robustly describe the image regions (or patches). The IKMs are further employed to build the BoVW model and fused with texture and color features to improve the semantic discriminative capability of each region and therefore achieve retrieval results semantically closer to human perception. Our extensive experiments show that SM-IKM outperforms state-of-the-art peer CBIR methods on two publicly available Caltech-101 and Wang datasets.

Unlike the SMFM method \cite{Bai2018}, which combines salient region detection and BoVW involving SIFT, LBP, and CH, the proposed SM-IKM employs IKM-based BoVW and fused features of IKMs, LBP of one color channel and the saliency map, and CH to capture the semantic meaning of each segmented region. The main contributions of this paper are as follows:
\begin{enumerate}
    \item Efficiently gaining more discriminative power by incorporating LBPs of the saliency map.
    \item Effectively gaining robustness against transformation, noise, and discretization error by employing IKMs.
    \item Efficiently reducing the BoVW construction time by producing compact discriminative features (e.g., IKMs).
    \item Balancing the runtime and memory complexity by providing the flexibility of specifying the length of the feature descriptors (i.e., the order of IKMs).
\end{enumerate}

The rest of this paper is organized as follows: Section \ref{sec:PrevWork} presents the related work which is the foundation of the proposed SM-IKM method. Section \ref{sec:ProposedMethod} explains the SM-IKM framework in detail. Section \ref{sec:ExpRes} summarizes the experimental results and compares the performance of SM-IKM with several state-of-the-art peer CBIR methods. Section \ref{sec:conclusion} concludes and presents directions for future work.

\section{Related Work} \label{sec:PrevWork}

In this section, we first briefly review the fundamentals of the two main components of the proposed SM-IKM method, namely, salient region detection and invariant Krawtchouk moments (IKMs). We then review the SMFM, a state-of-the-art peer CBIR method.

\subsection{Salient Region Detection}

Due to the enormous amount of data the human visual system receives, the brain automatically focuses on certain regions of the image \cite{Itti2000}. One of the goals of computer vision has always been to make the perception of computers close to the perception of humans. For this reason, many algorithms have been introduced to determine the prominent areas in an image. These methods are known as salient region detection algorithms and are also applied in various image retrieval frameworks \cite{Runxin2015}.

The output of a saliency region detection algorithm is a saliency map, which is a matrix of the same size as the input image. The value of each cell in the saliency map corresponds to the prominence of the corresponding pixel in the input image. This map can be further thresholded to separate the image’s foreground from its background.

The global contrast-based salient region detection algorithm \cite{Cheng2014} has been used in the proposed SM-IKM method to separate foreground from background. It uses a bottom-up saliency detection process to simultaneously evaluate global contrast differences and the spatially weighted coherence scores to extract full-resolution, high-quality saliency maps. It is simple, efficient, and multi-scale, making it suitable for our retrieval framework.

After the regions are approximately separated by applying a threshold on saliency maps, we extract and describe each region with appropriate features to convey complementary information. In addition, we describe each region differently by assigning different weights to their features.

\subsection{Invariant Krawtchouk Moments (IKMs)} \label{subsec:IKMs}
Krawtchouk moments (KMs) are calculated using Krawtchouk polynomials, which are discrete orthogonal polynomials associated with the binomial distribution. KMs can be mathematically transformed into translation, rotation, and scaling invariant features, called invariant Krawtchouk Moments (IKMs). The IKMs play a significant role in providing robust information against transformations, including translation, rotation, and scaling \cite{Yap2003}. IKMs are also highly discriminative and can be computed by approximating the image function with only six orders of Krawtchouk polynomials \cite{Sit2014}. We use the IKM features to provide vital invariance in the retrieval process.

The general moments $M_{pq}^{(f)}$ of an image  $f(x,y)$ is defined as follows:

\begin{equation}
    M_{pq}^{(f)} = \iint_D p_{pq}(x,y)\,f(x,y) \,dx\,dy,
\end{equation}
where $p$ and $q$ are non-negative integers indicating the order of moment $r=p+q$ and $p_{ij}(x,y)$ are polynomial basis functions defined on a compact support $D \subset R \times R$. Different kinds of moments are achieved depending on the polynomial basis employed \cite{Flusser2009}. Some of these moments can be mathematically adjusted to become resistant to transformations in $f(x,y)$, which are known as invariant moments.

By using $p_{kj}(x,y) = x^k \, y^j$ as the basis function, geometric moments are obtained by
\begin{equation} \label{eq:generalMoments}
    m_{pq}= \iint_{-\infty}^{+\infty} x^p y^q f(x,y) \, dx \, dy.
\end{equation}

Since an image is a discrete input function and its limits are the width and height of the image, the geometric moment can be rewritten as

\begin{equation} \label{eq:simpleGeomMom}
    m_{pq}=\sum_{x=1}^{h}\sum_{y=1}^{w} x^p y^q f(x,y),
\end{equation}
where $h$ and $w$ are the height and the width of the image, respectively.

It is proved that the set of rotation, scale, and translation invariant geometric moments \cite{Flusser2009} are computed by

\begin{equation} \label{eq:InvGeomMom}
    \begin{split}
        v_{nm} = {} & M_{00}^{-\dfrac{n+m}{2}-1}\sum_{x=1}^{N-1} \sum_{y=1}^{M-1} f(x,y) [(x-x_c)\cos{\theta} + (y-y_c)\sin{\theta}]^n \\
        & [-(x-x_c)\sin{\theta} + (y-y_c)\cos{\theta}]^m,
    \end{split}
\end{equation}
where $N$ and $M$ are the width and the height of an image, $n$ and $m$ are the orders of the moment, $x_c=m_{10}/m_{00}$ and $y_c=m_{01}/m_{00}$ are centroids of the object in an image, and $\tan{2\theta}=2\mu_{11}/(\mu_{20}+\mu{02})$.

Using Krawtchouk polynomials as the basis function of the moment leads to producing Krawtchouk moments. The $n$-th order of classic Krawtchouk polynomial \cite{Yap2003} is defined as

\begin{equation}
    K_n(x;p,N)=\sum_{k=0}^{N} a_{k,n,p}x^k = {}_{2}F_{1} (-n,-x;-N;1/p),
\end{equation}
where $x,n=0,1,2,…,N$ , $N>0$ , $p\in(0,1)$ and ${}_{2}F_{1}$ is the Hypergeometric function given as

\begin{equation}
    {}_{2}F_{1}(a,b;c;z) = \sum_{k=0}^{\infty} \dfrac{(a)_k (b)_k}{(c)_k} \dfrac{z^k}{k!}
\end{equation}
and $(a)_k$ is the Pochhammer symbol defined as

\begin{equation}
    (a)_k = a(a+1)...(a+k-1).
\end{equation}

The first three order of Krawtchouk polynomials \cite{Sit2014} are
\begin{equation}
    K_0=1,
\end{equation}
\begin{equation}
    K_1(x;p,N) = 1 - (\dfrac{1}{Np}) x,
\end{equation}
\begin{equation}
    K_2(x;p,N) = 1 - (\dfrac{2}{Np}+\dfrac{1}{N(N-1)p^2})x + (\dfrac{1}{N(N-1)p^2})x^2,
\end{equation}

The semicolons in the notations in this paper separate variables from parameters.

The range of values of the polynomials can fluctuate significantly with a slight change of order.To achieve numerical stability, normalization, and scaling, the square root of the weight can be employed to obtain the weighted Krawtchouk polynomials, which are defined as:

\begin{equation}
    \Bar{K}_n(x;p,N) = K_n(x;p,N) \sqrt{\dfrac{w(n;p,N)}{\rho(n;p,N)}},
\end{equation}
where
\begin{equation}
    w(x;p,N) = \binom{N}{x} p^x (1-p)^{N-x},
\end{equation}
and
\begin{equation}
    \rho(n;p,N) = (-1)^n \, (\dfrac{1-p}{p})^n \, \dfrac{n!}{(-N)_n}.
\end{equation}

Krawtchouk moments can also be written as a linear combination of geometric moments \cite{Sit2014}. Hence, we have a rotation, size, and translation invariant version of Krawtchouk moments by replacing the geometric moment with the invariant geometric moment.  By placing Krawtchouk polynomials as the basis function in Eq. (\ref{eq:generalMoments}), we define the weighted 2D Krawtchouk moments of order $n+m$ as

\begin{equation}
    \Bar{Q}_{nm} = \sum_{x=0}^{N}\sum_{y=0}^{M} f(x,y) \Bar{K}_n(x;p_x,N) \Bar{K}_m(y;p_y,M),
\end{equation}

\begin{figure}[ht]
\centering
\includegraphics[scale=0.37]{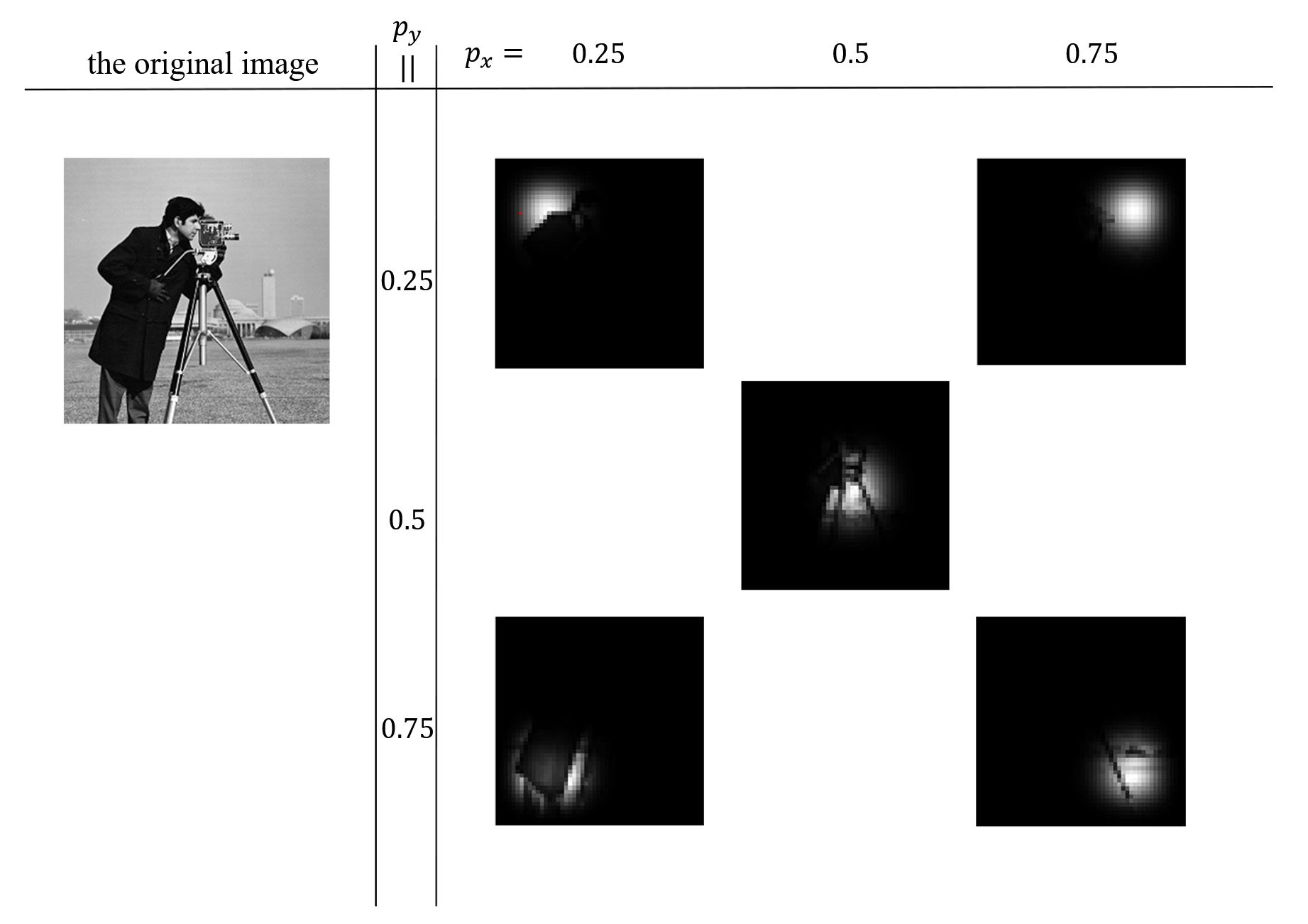}
\caption{\label{fig:cameramanWeights} The focus zones of the multi-order IKM description on the Cameraman image, whose size is resized from 512x512 to 52x52 to better demonstrate the effect of the weight function on the small patches.}
\end{figure}

The values of $p_x$ and $p_y$ directly determine the center of focus. When changing this pair of values, the descriptor focuses on any desired area, known as region-of-interest \cite{Yap2003}. Fig. \ref{fig:cameramanWeights} demonstrates the focus zones of the IKMs with different order pairs of ($p_x$, $p_y$) for a sample image and shows the effect of different values of $p_x$ and $p_y$ on region-of-interest.

To obtain the 2D Krawtchouk moment, the weight function for an input image of size $N \times M$ is rewritten as
\begin{equation} \label{eq:2dWeightFunc}
    w(x,y;p_x,p_y) = \sqrt{w(x;p_x,N-1)w(y;p_y,M-1)}.
\end{equation}

Pixel-wise multiplication of the input image and the weight matrix yields the weighted input as
\begin{equation}
    \Tilde{f}(x,y) = w(x,y;p_x,p_y) f(x,y).
\end{equation}

The weighted Krawtchouk moment for a 2D input is then defined as
\begin{equation}
    \begin{split}
        \Bar{Q}_{nm} = {}  \sum_{x=0}^{N}\sum_{y=0}^{M} \Tilde{f}(x,y) \Bar{K}_n(x;p_x,N) \Bar{K}_m(y;p_y,M) \\
         = {} [\rho(n;p_x;N-1) \rho(m;p_y,M-1)]^{-\dfrac{1}{2}}\\
         \sum_{i=0}^{n} \sum_{j=0}^{m} a_{i,n,p_x,N-1} a_{j,m,p_y,M-1} m_{ij}.
    \end{split}
\end{equation}
where $m_{ij}$ is the geometric moment defined in Eq. (\ref{eq:simpleGeomMom}). Hence, weighted Krawtchouk moments are computable using geometric moments. When replacing them with invariant geometric moments defined in Eq. (\ref{eq:InvGeomMom}), the robustness against translation, scaling, and rotation is achievable.

It should be noted that some orders of IKMs have constant values regardless of the input. Therefore, using all the possible $N \times M$ orders is not the most efficient solution. Sit and Kihara \cite{Sit2014} suggest that the following specific set of orders can be used:

\begin{equation} \label{eq:krawFeatVec}
    V = [\Tilde{Q}_{02},\Tilde{Q}_{20},\Tilde{Q}_{12},\Tilde{Q}_{21},\Tilde{Q}_{30},\Tilde{Q}_{03} ]
\end{equation}

We use the six IKM features from Eq. (\ref{eq:krawFeatVec}) to describe both foreground and background of the image.

\subsection{Saliency-based Multi-Feature Modeling (SMFM) Method} \label{sec:SMFM}

Bai et al. \cite{Bai2018} propose the SMFM framework, which uses the global contrast-based salient region detection method \cite{Cheng2014} to separate the foreground and background regions of each image. They then extract a series of features, including histograms of hue and saturation channels in hue, saturation, and value (HSV) color space, as well as SIFT and LBPs for each region. To this end, the input image is converted to HSV and gray-scale representation. The histogram of intensity values of the H channel (i.e., $H_h$), the histogram of intensity values of the S channel (i.e., $H_s$), and local binary patterns ($LBP$) \cite{Ojala2002} of the V channel (i.e., $LBP_v$) are computed for both foreground and background regions. In addition, the SIFT descriptor is computed on the gray-scale representation of the image. Its portion in the foreground region is transformed into a histogram of SIFT word occurrences ($SIFT_g$) in the BoVW, which is constructed using the K-means clustering algorithm on the global SIFT features and functions as the vocabulary obtained from the gray-scale representation of an image. Each image has a seven-part feature (three for the background and four for the foreground), which is formulated as

\begin{equation} \label{eq:SMFMFeats}
    F =
\begin{cases}
    F_f = \{H_h, H_s, LBP_v, SIFT_g\},\\
    F_b = \{H_h, H_s, LBP_v\},
\end{cases}
\end{equation}
where $F_f$ and $F_b$ are the features extracted from foreground and background, respectively.

The SMFM uses the color and texture distribution to represent both foreground and background regions since they carry valuable information for the recognition objective. It adds SIFT occurrences features in the foreground region to provide an invariant description of the foreground, which tends to contain the main object. It finally uses Chi-Square as the similarity measurement to calculate the distance between each feature and uses the z-score normalization to compute the normalized distance by:

\begin{equation} \label{eq:chiDist}
    DS_A^N(Q,I_i) = \dfrac{DS_A(Q,I_i) - \mu_{AQ}}{\sigma_{AQ}},
\end{equation}
where $\mu_{AQ}$ and $\sigma_AQ$ are the mean value and the standard deviation of the distances set ${DS_A(Q,I_i)}$, respectively.

The normalized distances are then fused according to a predefined weight for each feature. These weights are empirically specified. Specifically, the weight of 1’s is chosen for $H_s$ and $H_h$ of both background and foreground. The background weight of $LBP_v$ is set to be 1 and the foreground weight of $LBP_v$ is set to be 3. A weight of 1 is chosen for $SIFT_g$, whose length is 128.

\section{Proposed Method} \label{sec:ProposedMethod}

Fig. 2 presents the data pipeline of the proposed saliency map-based image retrieval using invariant Krawtchouk moments (SM-IKM). SM-IKM improves SMFM \cite{Bai2018} by efficiently acquiring the IKMs to replace SIFT features and enhancing the usage of the saliency map by using LBPs of the entire saliency map. All the improved parts are shown in the highlighted boxes. In the following, we explain each improved part in detail.

\subsection{Saliency Map and LBP Feature Extraction}
We employ the global contrast-based salient region detection method \cite{Cheng2014} to obtain a full-resolution saliency map. SaliencyCut iteratively refines the salient regions in the saliency map to achieve high-quality unsupervised salient object segmentation to separate foreground from background. Unlike SMFM, which discards the saliency map after separating the background and foreground regions, SM-IKM uses the saliency map to get a very compelling feature, namely, the LBP of the saliency map, to represent the texture of an image.

\subsection{Patch Extraction Using SIFT and Invariant Krawtchouk Descriptor}
Although invariant Krawtchouk moments (IKMs) can approximate the entire image, we use them as a local descriptor to describe the patches extracted from the image. The process of patch extraction is carried out by locating key points using SIFT. A patch is then cropped around each key point of interest. The IKMs, whose length is six, are extracted from each patch by Eq. (\ref{eq:krawFeatVec}). The resultant IKMs from all patches are then stored to create the vocabulary by executing the K-means algorithm. When the vocabulary is computed, the histogram of occurrences of each vocabulary is obtained for each image to represent its semantic contents.

Since IKMs are compact with a length of six, their use in the k-means clustering algorithm to generate the BoVW leads to significantly reduced time to efficiently construct a BoVW model while putting more focus on the prominence pixels in the input image \cite{Runxin2015}.

\subsection{Multi-order SM-IKM and Its Two Variants}
Similar to SMFM, the proposed SM-IKM acquires multiple features representing each image. Unlike SMFM, which uses SIFT as the core feature responsible for invariance to transformations, SM-IKM applies the IKMs due to its high compactness, high discriminative power, and robustness against transformations. It also transforms an image into the HSV color representation to obtain the color histograms of H and S channels to provide the color distribution of the image. In addition, the LBP of channel V is used to obtain texture information. LBP of the saliency map provides beneficial information using the prominence pixels in the input. In summary, the proposed SM-IKM uses eight features to represent an image from multiple perspectives: three features from the background, four from the foreground, and one feature from the saliency map of the entire image. The structure of the final weighted features is as follows:

\begin{equation} \label{eq:SMKMFeats}
    F =
\begin{cases}
    F_f = \{2 (H_h), 2 (H_s), 3 (LBP_v), 1.5 (Kraw_g)\},\\
    F_b = \{1 (H_h), 1 (H_s), 2 (LBP_v)\},\\
    F_{sm} = \{1.5 (LBP_{sm}),\},
\end{cases}
\end{equation}
where $F_f$, $F_b$, and $F_{sm}$ respectively, the features extracted from the foreground, background, and saliency map and $H_h$ and $H_s$ are color histograms of channels H and S, $LBP_v$ and $LBP_{sm}$ are respectively the LBPs of the channel V and the saliency map, and $Kraw_g$ is the histogram of foreground IKMs occurrences in the gray-scale representation of the image based on computed BoVW. Finally, the distances are feature-wise computed, normalized, and fused into one final value by Eq. (\ref{eq:chiDist}). The obtained value gives the similarity between the two images and is used to determine the most relevant and similar images to a query. Specifically, a higher value represents more semantic similarity between the two images.

The parameters $p_x$ and $p_y$ in Eq. (\ref{eq:2dWeightFunc}) determine the center of attention in the patch or the whole image. In the proposed method, we use multi-order IKMs (i.e., five pairs of orders of $p_x$ and $p_y$) to describe each patch using different attention centers. These pairs of orders are (0.25, 0.25), (0.25, 0.75), (0.5, 0.5) (0.75, 0.25), and (0.75, 0.75), where attention centers are in the upper-left corner, the upper-right corner, the center, the lower-left corner, and the lower-right corner, respectively.  In other words, we describe each patch five times, each time with attention slightly closer to one corner of the image, and the center of attention is the middle of the five patches, as shown in Fig. \ref{fig:cameramanWeights}. We call this feature multi-order and name the corresponding SM-IKM as multi-order SM-IKM. The resulted vector for each pair of orders is concatenated to form a vector of size 30 to represent training and testing images.

We also developed two variants of the proposed multi-order SM-IKM. The first variant sets both $p_x$ and $p_y$ to be equal to 0.5,  leading to a descriptor of length six that focuses more on the center area of an image or a center patch than its surroundings. We name it single-order SM-IKM. The second variant is to remove the LBP features of the salient map from the first variant. We name it single-order SM-IKM without $LBP_{sm}$.

\begin{figure*}[ht] \label{fig:krawFlowDiag}
\centering
\includegraphics[scale=1]{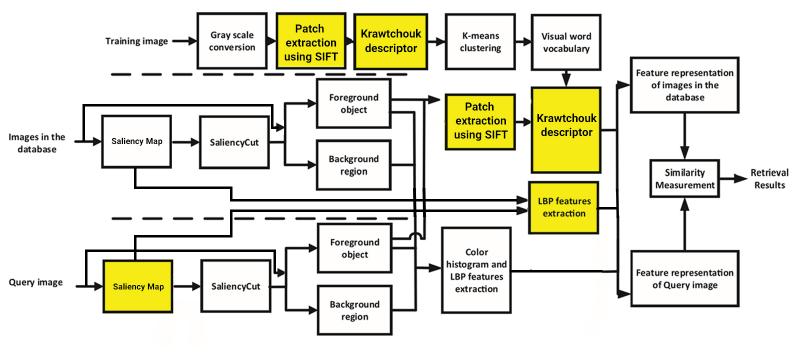}
\caption{Data pipeline of the proposed SM-IKM method.}
\end{figure*}

\section{Experimental Results} \label{sec:ExpRes}
This section presents experiments to evaluate the performance of the proposed SM-IKM method and several state-of-the-art peer methods. The code is implemented in MATLAB and executed on a system with 16 GB of RAM and Core i7-6560U. The code for our framework is available on \href{https://github.com/arnejad/SMIKM}{https://github.com/arnejad/SMIKM} repository.

Similar to the 2DKD library \cite{Deville2020} published in 2020, our implementation computes the six orders of moments as specified in Eq. (\ref{eq:krawFeatVec}). This library also computes the IKMs and the rest of the feasible orders.

\subsection{Datasets}
We evaluate the proposed SM-IKM framework by conducting experiments on two publicly available datasets: Wang \cite{Li2003} and Caltech-101 \cite{Fei2006}.

The Wang dataset is comprised of 1000 images in 10 categories, including African people, beach, building, bus, dinosaur, elephant, flower, horse, mountain, and dish. This dataset is a subset of the Flickr database. Each image has a label that belongs to one of the classes. All of the images in the Wang dataset contribute to both training and testing.

Caltech-101 is a dataset containing about 9000 real-world image samples in 101 classes, including camera, plane, elephant, football, chair, etc. The number of images in each class varies from 40 to 800, where an average of approximately 50 images in each class. In our experiments, gray-scale images are discarded except for the car class, which contains only gray-scale images. We select 300 images used in \cite{Zhang2017} for training and use 680 images from the four categories of leopard, plane, sailboat, and pelican for testing.

\subsection{Comparison of Several State-of-the-art CBIR methods}

We conduct extensive experiments to find the best local texture patterns that suit SM-IKM. To this end, we implement and evaluate 27 local descriptors summarized in \cite{Turan2018} and report the top eight descriptors in terms of mean average precision (mAP). These top eight descriptors include local binary patterns (LBP) \cite{Ojala2002}, improved Weber binary coding (IWBC) \cite{Yang2016}, local directional number patterns (LDN) \cite{Rivera2012}, local monotonic patterns (LMP) \cite{Mohammad2011}, local arc patterns (LAP) \cite{Islam2014}, gradient directional patterns (GDP) \cite{Ahmed2012}, local gradient increasing patterns (LGIP) \cite{Zhou2012}, and local phase quantization (LPQ) \cite{Ojansivu2008}. It should be noted that we examine different configurations of the eight aforementioned methods and report the results obtained in the best setting. For example, we extract LDNs using three kinds of filters (e.g., Sobel, Gaussian, and Krisch), each of which has three kinds of window sizes (3x3, 5x5, and 7x7). We employ three different scales to extract IWBC features. Table \ref{tab:tpComp} reports the class-specific mAP of single-order SM-IKM on the Wang dataset, where LBP features are replaced with each of the top eight descriptors. It clearly shows that the LBP features achieve the best mAP on the Wang dataset and the best retrieval accuracy on the four categories (i.e., building, flower, mountain, and food). Therefore, LBP features are selected as the final features in the proposed SM-IKM method. To ensure a fair comparison, the same vocabulary is used for all the eight variants of the single-order SM-IKM. We also set the patch size to 30x30 and set the vocabulary size to 100.

\begin{table*}[!t]
\caption{Comparison of the mAP of the proposed single-order SM-IKM method on the Wang dataset, where each of eight top local texture pattern descriptors replaces the LBPs.}
    \label{tab:tpComp}
    \centering
    \begin{tabularx}{\linewidth}{|X|c|c|c|c|c|c|c|c|}
    \hline
        Class & LBP \cite{Ojala2002} & IWBC \cite{Yang2016} & LDN \cite{Rivera2012} & LMP \cite{Mohammad2011} & LAP \cite{Islam2014}  & GDP \cite{Ahmed2012} & LGIP \cite{Zhou2012} & LPQ \cite{Ojansivu2008} \\
    \hline
        African     & 69.41\% & 67.59\% & 67.57\% & 67.69\% & \textbf{69.71}\% & 65.73\% & 68.53\% & 64.31\%  \\
        beach       & 50.28\% & 48.91\% & \textbf{51.73}\% & 49.20\% & 50.69\% & 47.04\% & 48.10\% & 49.09\%  \\
        building    & \textbf{63.29}\% & 60.75\% & 56.76\% & 61.41\% & 60.88\% & 58.29\% & 61.28\% & 57.23\%  \\
        bus         & 87.55\% & 87.52\% & \textbf{90.03}\% & 86.90\% & 87.27\% & 87.01\% & 86.99\% & 86.16\%  \\
        dinosaur    & 99.24\% & 99.27\% & 97.06\% & \textbf{99.42}\% & 99.40\% & 99.08\% & 99.29\% & 98.41\%  \\
        elephant    & 57.87\% & \textbf{60.29}\% & 58.03\% & 60.03\% & 59.81\% & 57.14\% & 59.30\% & 60.22\%  \\
        flower      & \textbf{82.72}\% & 81.78\% & 74.62\% & 83.30\% & 82.17\% & 81.31\% & 79.43\% & 79.74\%  \\
        horse       & 91.69\% & \textbf{92.15}\% & 91.00\% & 92.07\% & 91.79\% & 91.18\% & 91.97\% & 91.35\%  \\
        mountain    & \textbf{48.50}\% & 47.71\% & 44.88\% & 48.15\% & 48.01\% & 46.62\% & 47.70\% & 46.87\%  \\
        food        & \textbf{74.26}\% & 72.04\% & 65.47\% & 72.09\% & 73.24\% & 70.21\% & 73.05\% & 67.25\%  \\
        \hline
        mean        & \textbf{72.48}\% & 71.80\% & 69.72\% & 72.09\% & 72.30\% & 70.36\% & 71.56\% & 70.06\%  \\
    \hline
    \end{tabularx}
\end{table*}

Table \ref{tab:classSpecWang} summarizes the performance of three compared methods in terms of the retrieval accuracy for each class and the mAP of all the ten classes of the Wang dataset. The three compared methods include SMFM, single-order SM-IKM (the first variant of SMFM), and multi-order SMFM (proposed SMFM). It clearly shows that the proposed SM-IKM method and its variant outperform its peer SMFM method. Specifically, single-order and multi-order SM-IKM methods improve the mAP of SMFM by 4.23\% and 4.43\%, respectively. The multi-order SM-IKM method achieves the best retrieval accuracy in five classes: bus, dinosaur, elephant, flower, and horse. The single-order SM-IKM achieves the best retrieval accuracy in the remaining five classes.

\begin{table}[]
    \caption{Comparison of the mAP of the proposed single-order and multi-order SM-IKM methods and the peer SMFM method for each image category of the Wang dataset.}
    \label{tab:classSpecWang}
    \centering
    \begin{tabularx}{\linewidth}{|X|c|c|c|}
    \hline
        Class       & SMFM & \multicolumn{2}{c|}{SM-IKM}\\
           &                & single-order     & multi-order \\
    \hline
        African     & 69.08\%   & 69.41\%   & 68.06\% \\
        beach       & 45.69\%   & 50.28\%   & 49.93\% \\
        building    & 55.19\%   & 63.29\%   & 60.94\% \\
        bus         & 85.32\%   & 87.55\%   & 90.04\% \\
        dinosaur    & 99.20\%   & 99.24\%   & 99.37\% \\
        elephant    & 54.71\%   & 57.87\%   & 60.75\% \\
        flower      & 81.00\%   & 82.72\%   & 84.29\% \\
        horse       & 91.47\%   & 91.69\%   & 92.35\% \\
        mountain    & 45.39\%   & 48.50\%   & 47.81\% \\
        food        & 68.35\%   & 74.26\%   & 72.70\% \\
        \hline
        mean        & 69.54\%   & 72.48\%   & 72.62\% \\
    \hline
    \end{tabularx}
\end{table}

Table \ref{tab:WangRes} compares the performance of the proposed SM-IKM method, its two variants (single-order SM-IKM and single-order SM-IKM without $LBP_{sm}$), SMFM, and two other retrieval methods proposed in \cite{Singh2018} and \cite{Dubey2016} on the Wang dataset. Table \ref{tab:CalRes} compares the performance of the proposed SM-IKM method, its two variants, SMFM, and two other retrieval methods proposed in \cite{Zhang2017} on 680 images from four classes of the Caltech-101 dataset. To ensure a fair comparison of the three SM-IKM methods, we construct the same vocabulary in BoVW for each of the two datasets using their respective training images.

Both Tables \ref{tab:WangRes} and \ref{tab:CalRes} show that the proposed multi-order SM-IKM achieves the best mAP of 72.62\% and 93.96\% on the Wang dataset and the Caltech-101 dataset, respectively. However, none of the other methods consistently achieves the second-best performance on the two datasets. For example, the proposed single-order SM-IKM ranks second on the Wang dataset and third on the Caltech-101 dataset. The proposed single-order SM-IKM without $LBP_{sm}$ ranks third on the Wang dataset and fourth on the Caltech-101 dataset. The peer method SMFM ranks fourth on the Wang dataset and second on the Caltech-101 dataset. The proposed multi-order SM-IKM improves the mAP of the second-best retrieval method (single-order SM-IKM) for the Wang dataset by 0.19\% and the mAP of the second-best retrieval method (SMFM) for the Caltech-101 dataset by 0.85\%. For the Wang dataset, three proposed SM-IKM methods outperform the three compared peer methods. Specifically, multi-order SM-IKM, single-order SM-IKM, and single-order SM-IKM without $LBP_{sm}$ respectively improve the mAP of SMFM (the best peer retrieval method) by 4.43\%, 4.23\%, and 0.75\%. For the Caltech-101 dataset, SMFM outperforms two variant systems. However, the same three proposed methods improve the mAP of the extended saliency-based method (the second-best peer retrieval method) by 5.24\%, 4.21\%, and 2.18\%, respectively.

\begin{table}[t]
    \caption{ompares the performance of the proposed SM-IKM method, its two variants (single-order SM-IKM and single-order SM-IKM without $LBP_{sm}$), SMFM, and two other retrieval methods proposed in cite{Singh2018} and \cite{Dubey2016} on the Wang dataset.}
    \label{tab:WangRes}
    \centering
    \begin{tabularx}{\linewidth}{|X|c|}
    \hline
        Method                              & Average mAP           \\
    \hline
        LBPC+LBPH+CH\cite{Singh2018}        & 65.16\%           \\
        MDLBP \cite{Dubey2016}              & 60.82\%           \\
        SMFM \cite{Bai2018}                 & 69.54\%           \\
        SM-IKM (single-order without $LBP_{sm}$)   & 70.06\%           \\
        SM-IKM (single-order)                 & 72.48\%           \\
        SM-IKM (multi-order)                  & \textbf{72.62\%}  \\
    \hline
    \end{tabularx}
\end{table}

\begin{table}[t]
    \caption{Comparison of the mAP of the proposed SM-IKM, its two variants, and three peer methods on the Caltech-101 dataset.}
    \label{tab:CalRes}
    \centering
    \begin{tabularx}{\linewidth}{|X|c|}
    \hline
        Method                                          & Average mAP      \\
    \hline
        Saliency-based method \cite{Zhang2017}          & 88.35\%          \\
        Extended saliency-based method \cite{Zhang2017} & 89.28\%          \\
        SMFM\cite{Bai2018}                              & 93.17\%          \\
        SM-IKM (single-order without $LBP_{sm}$)               & 91.23\%          \\
        SM-IKM (single-order)                             & 93.04\%          \\
        SM-IKM (multi-order)                              & \textbf{93.96\%} \\
    \hline
    \end{tabularx}
\end{table}

The retrieval results on both datasets show that the proposed SM-IKM performs the best, and its first variant outperforms its second variant. Here are a few observations and conclusions:
\begin{enumerate}
    \item Multi-order SM-IKM achieves a performance gain of 0.19\% on the Wang dataset and 0.99\% on the Caltech-101 dataset compared with single-order SM-IKM. We can conclude that incorporating multiple orders of IKMs to consider the invariant features at multiple focus zone tend to boost retrieval accuracy.
    \item Single-order SM-IKM achieves a significant performance gain of 3.45\% on the Wang dataset and 1.98\% on the Caltech-101 dataset when compared with single-order SM-IKM without $LBP_{sm}$. We can conclude that incorporating $LBP_{sm}$ to represent the texture information of the salient map significantly boosts retrieval accuracy.
    \item Single-order SM-IKM without $LBP_{sm}$ achieves a performance gain of 0.75\% on the Wang dataset and a performance loss of 2.08\% on the Caltech-101 dataset when compared with SMFM. We can conclude that incorporating the compact IKMs to replace SIFT achieves robustness against the transformation without significantly compromising retrieval accuracy. In addition, it takes much less time to construct the vocabulary than similar methods that use the BoVW model since the length of IKMs is six for single-order SM-IKM and 30 for multi-order SM-IKM, and the length of SIFT is 128.
    \item SMFM achieves a performance gain of 6.72\% on the Wang dataset when compared with LBPC+LBPH+CH and achieves a performance gain of 4.36\% on the Caltech-101 dataset when compared with the extended saliency-based method. We can conclude that incorporating the transformation invariant SIFT features and other histogram-based features boosts retrieval accuracy.
\end{enumerate}

\subsection{Comparison of Computational Runtime}
\begin{figure}[]
\centering
\includegraphics[scale=0.6]{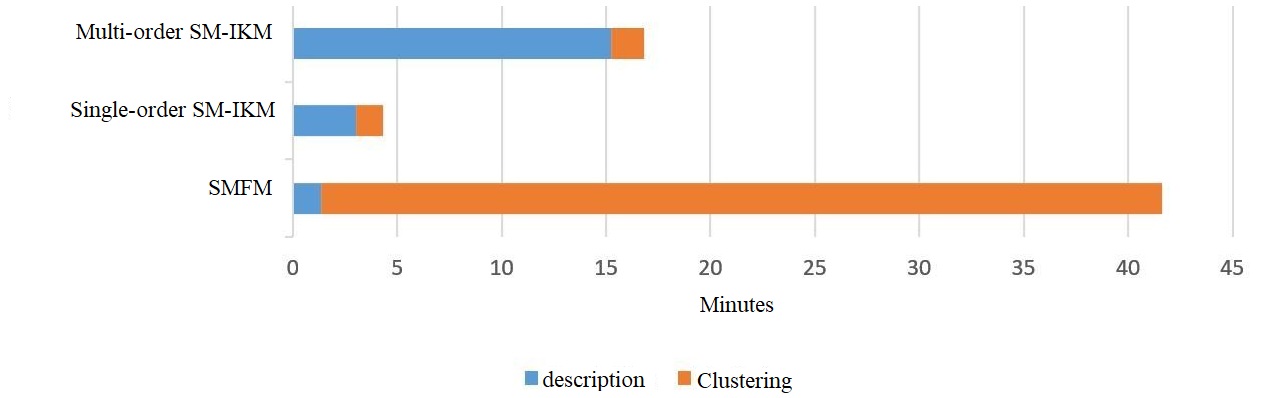}
\caption{\label{fig:runtime}Comparison of the retrieval time (in minutes) of the top three methods for all 1000 images in the Wang dataset, which includes the feature description computation time (shown in blue) and the clustering time (shown in orange) into 100 vocabularies of the BoVM model.}
\end{figure}

Fig. \ref{fig:runtime} compares the retrieval time (in minutes) of the top three methods for all 1000 images in the Wang dataset, which is calculated by averaging the computational time to obtain retrieval response for all the images in the
Wang dataset. This retrieval time includes the time to extract feature description and the time to construct the BoVM model (i.e., use the extracted features to cluster 1000 images of the Wang dataset into 100 clusters). It shows that the runtime needed to calculate SIFT features is less than the runtime needed to calculate the IKMs. However, the overall runtime (extracting features and clustering) of SM-IKM variants, especially the single-order SM-IKM, is significantly lower than the overall runtime of the SMFM method since the low-dimension feature vector reduces the clustering time to build the BoVM model. In summary, it takes significantly less time for SM-IKM and its variants than SMFM and its similar methods to retrieve an image.

\section{Conclusion} \label{sec:conclusion}

This paper proposes an efficient saliency map-based image retrieval using invariant Krawtchouk moments (SM-IKM) to achieve fast and accurate content-based image retrieval results. First, the global contrast-based salient region detection algorithm is employed to separate foreground from background. Second, multiple features are extracted from the foreground, background, and the saliency map. Specifically, color histograms and LBPs are extracted to provide color and texture properties for both foreground and background. Invariant Krawtchouk moments (IKMs) are extracted from the foreground to gain robustness against the possible transformations in the main object since they do not carry discretization error and can be used as local descriptors. LBPs of the saliency map is also extracted to represent images from another holistic perspective. Third, the BoVW model is constructed on the compact IKMs, whose length can be 6 for single-order SM-IKM or 30 for multi-order SM-IKM. Fourth, the Chi-square-based similarity metric is employed to measure the similarity between the query image and the images in the database. The similarity scores are also normalized and fused using a pooling strategy to obtain the final scores. Our extensive experimental results show that the proposed SM-IKM method outperforms the state-of-the-art CBIR methods on two publicly available datasets. Its variants also achieve comparable retrieval accuracy to the peer SMFM method.

\bibliographystyle{unsrt}  
\bibliography{references}  

\end{document}